\documentclass{article}

\usepackage[preprint]{neurips_2026}
\usepackage[utf8]{inputenc}
\usepackage[T1]{fontenc}
\usepackage{float}
\usepackage{amsmath,amssymb,amsfonts}
\usepackage{booktabs}
\usepackage{graphicx}
\usepackage{xcolor}
\usepackage{microtype}
\usepackage{nicefrac}
\usepackage{url}
\usepackage{csquotes}

\usepackage{tikz}
\usetikzlibrary{arrows.meta,positioning,shapes,calc,fit}
\usepackage{algorithm}
\usepackage{algorithmic}
\graphicspath{{./images/}}

\usepackage[colorlinks=true,
            linkcolor=blue,
            citecolor=blue,
            urlcolor=blue]{hyperref}

\title{Safe Continual Reinforcement Learning under Nonstationarity via Adaptive Safety Constraints}

\author{%
Timofey Tomashevskiy\\
McMaster Centre for Software Certification\\
Department of Computing and Software\\
McMaster University, Canada\\
\texttt{tomashet@mcmaster.ca}
}

\begin{document}

\maketitle

\begin{abstract}
Safe continual reinforcement learning in nonstationary environments requires constraints that remain aligned with changing dynamics, costs, and operating context. Fixed safety specifications, fixed policy-level safety criteria, and reactive shields can become obsolete or intervene too late. We propose LILAC+, a predictive safety layer that constructs three adaptive constraint mechanisms using detected context and context forecasts: context-based (CB) constraints adapt existing safety constraints to current and predicted environments, adaptation-speed (AS) constraints compare the speed required for safe adaptation with the agent's estimated adaptation capacity and tighten behavior when safe adaptation may be too slow, and soft-to-hard (SH) constraints convert remaining safety budget into local enforceable thresholds. These mechanisms support both policy-level regularization and action-level shielding. Experiments in the \texttt{highway-env} \texttt{merge-v0} task, across stationary, seen nonstationary, and unseen nonstationary conditions, show fewer safety violations than unconstrained and fixed-constraint baselines while maintaining competitive reward. The results demonstrate a practical mechanism for predictive safety adaptation, while the claims remain conditional on context-estimation quality and feasible enforcement.
\end{abstract}
\section{Introduction}

Standard safe reinforcement learning formulations specify safety through fixed costs, constraints, or admissible sets defined for a given task or environment~\citep{garcia2015comprehensive,gu2022review}. Under nonstationarity, such constraints can fail in two ways. First, a policy may satisfy an outdated constraint that no longer captures safety in the changed environment. Second, even when the appropriate constraint can be updated, environmental change may leave too little time to implement the revised constraints through policy adaptation or safe action selection. Thus, safety failures may arise from two forms of mismatch: obsolete constraints following an environmental shift and delayed implementation of revised constraints under rapid environmental change.

\begin{figure}[t]
    \centering
    \includegraphics[width=0.56\linewidth]{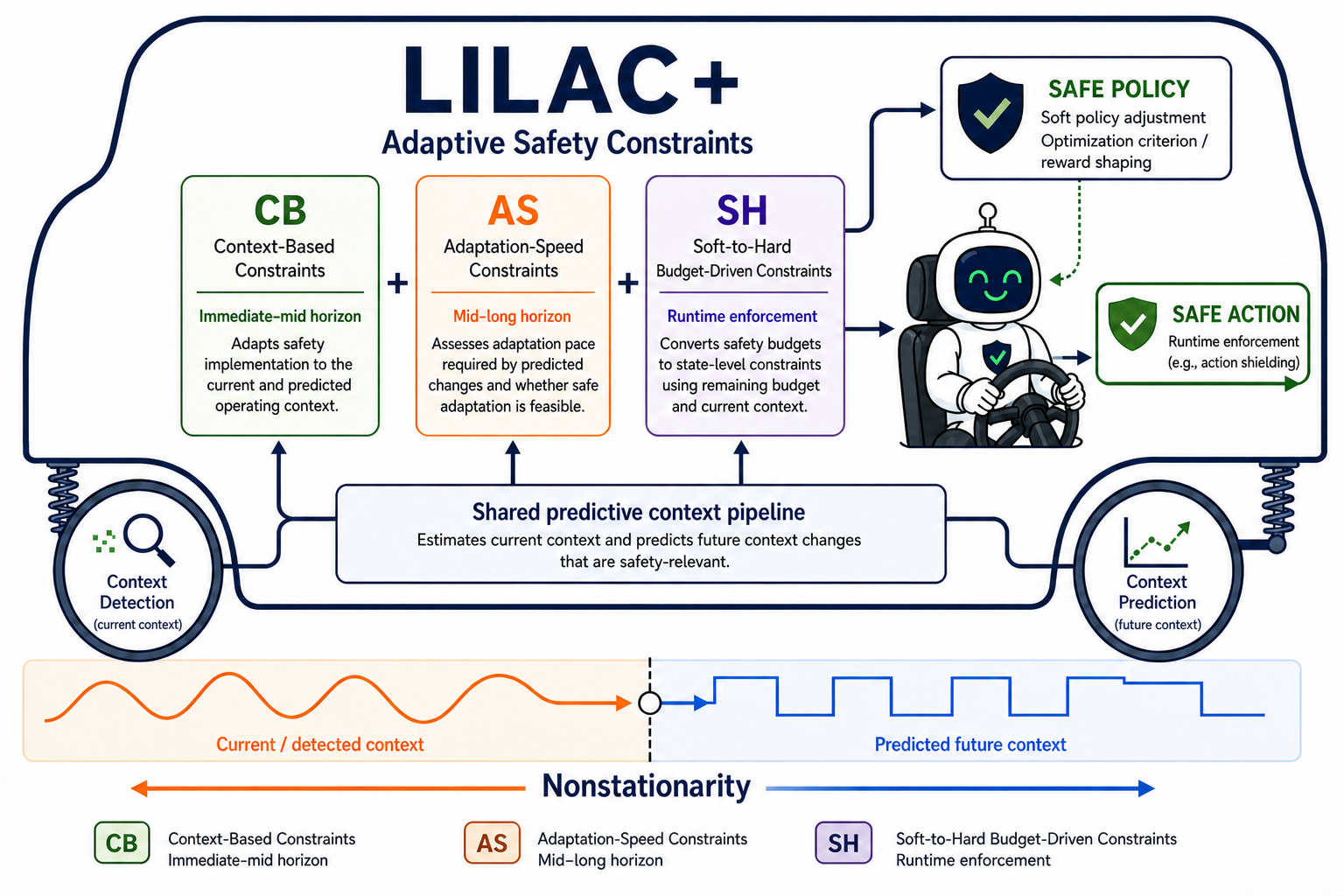}
    \caption{
Overview of LILAC+. A shared predictive context pipeline estimates the current context and predicts safety-relevant future context changes under nonstationarity. The current and predicted context parameterize three adaptive constraint types: context-based constraints (CB), adaptation-speed constraints (AS), and soft-to-hard budget-driven constraints (SH). CB supports immediate-to-mid-horizon adjustment, AS supports mid-to-long-horizon safety assessment by comparing the speed required for safe adaptation with the agent's estimated adaptation capacity, and SH converts safety budgets into state-level constraints for runtime enforcement. The resulting constraints support both soft policy adjustment and safe action selection.
}
\label{fig:intro_framework}
\end{figure}

This issue is especially important in safety-critical domains such as autonomous driving, robotic control, and adaptive cyber-physical systems, where unsafe exploration can have severe consequences. The critical risk is often the transient period in which the agent detects, interprets, and adapts to environmental change. Figure~\ref{fig:intro_framework} illustrates our approach.

We argue that safety in nonstationary reinforcement learning should be formulated as \emph{context-dependent constraint adaptation}. High-level safety goals, such as avoiding collisions or limiting cumulative risk, may remain stable, but the concrete constraints needed to satisfy those goals may change with context. LILAC+ uses context detection and context prediction to adapt safety constraints proactively before anticipated changes increase violation risk. Predicted context changes affect safety in three complementary ways: they determine which constraints are appropriate, whether the agent can adapt quickly enough, and how much remaining safety budget can be allocated to future actions.

Building on this perspective, we propose LILAC+, a predictive-context safety layer for continual reinforcement learning under nonstationarity. The key idea is that context should not only guide policy optimization in unconstrained settings~\citep{xie2020deep}, but should also parameterize the safety constraints used to shape learning and restrict execution. Unlike fixed specifications or predefined action filters~\citep{alshiekh2018safe,jansen2020safe,carr2023safe}, LILAC+ updates constraints using current and predicted context before anticipated changes increase violation risk.

LILAC+ uses three complementary constraint mechanisms. CB adapts existing safety constraints to current and predicted environments by comparing the predicted context with known contexts. AS compares the speed required for safe adaptation with the agent's estimated adaptation capacity and treats insufficient adaptation capacity as a safety signal. SH converts cumulative safety budgets into local enforceable thresholds. Together, these mechanisms adapt safety constraints for both policy-level learning and action-level enforcement.

We evaluate LILAC+ in the \texttt{highway-env} \texttt{merge-v0} driving task~\citep{leurent2018highwayenv} under stationary, seen nonstationary, and held-out nonstationary conditions. This controlled domain tests the proposed mechanisms but does not establish domain-general safety. The results indicate fewer safety violations than unconstrained and fixed-constraint baselines while maintaining competitive reward.
\paragraph{Contributions.}
This work makes four contributions.
\textbf{(i)} We formulate safe continual reinforcement learning under nonstationarity as proactive, context-dependent constraint adaptation anchored to stable high-level safety goals.
\textbf{(ii)} We develop three mechanisms: CB adapts existing safety constraints to current and predicted environments, AS identifies future state regions where the speed required for safe adaptation may exceed the agent's estimated adaptation capacity, and SH converts cumulative safety budgets into local enforceable thresholds.
\textbf{(iii)} We instantiate these mechanisms in a predictive safety layer with concrete domain-specific definitions of $D$, $F$, $v_t^{\mathrm{cap}}$, and $\phi$ for the evaluated environment.
\textbf{(iv)} We evaluate the mechanisms in \texttt{merge-v0}, showing reduced violations and cumulative safety cost with competitive reward.
\section{Background and Problem Formulation} \label{sec:background}

\paragraph{Nonstationary reinforcement learning.}
We consider reinforcement learning in environments whose dynamics, rewards, or safety costs may change over time. At time $t$, the agent observes state $s_t\in\mathcal{S}$, selects action $a_t\in\mathcal{A}$, receives reward $r_t$, and incurs safety cost $c_t\ge 0$. Nonstationarity is represented by a time-varying context $\kappa_t\in\mathcal{K}$ that captures safety-relevant operating conditions. The context may include observable variables as well as latent factors inferred from interaction data. We write
\begin{equation}
    s_{t+1}\sim P_{\kappa_t}(\cdot\mid s_t,a_t),
    \qquad
    r_t=r_{\kappa_t}(s_t,a_t),
    \qquad
    c_t=c_{\kappa_t}(s_t,a_t),
    \label{eq:nonstationary_dynamics}
\end{equation}
where changes in $\kappa_t$ may affect dynamics, rewards, costs, or constraint validity. The true context need not be directly observed; the agent may condition its behavior on an estimate $\hat{\kappa}_t$, yielding a context-conditioned policy $\pi(a_t\mid s_t,\hat{\kappa}_t)$.

\paragraph{Context detection and prediction.}
The agent maintains an interaction history
\begin{equation}
    \mathcal{H}_t=(s_0,a_0,r_0,c_0,\ldots,s_{t-1},a_{t-1},r_{t-1},c_{t-1},s_t),
\end{equation}
and uses context detection to estimate the current context,
\begin{equation}
    \hat{\kappa}_t=D(\mathcal{H}_t).
    \label{eq:context_detection}
\end{equation}
A context predictor estimates future safety-relevant context over horizon $H$,
\begin{equation}
    \hat{\kappa}_{t+1:t+H}=F(\mathcal{H}_t,\hat{\kappa}_t).
    \label{eq:context_prediction}
\end{equation}
We use \emph{context} broadly to denote information that affects dynamics, rewards, costs, or the appropriate implementation of safety constraints.

\paragraph{Two safety requirements.}
Safe continual reinforcement learning must address two related but distinct requirements. First, the learned policy should remain safe in a cumulative or probabilistic sense while maximizing return. Second, when a hard safety specification is active, the executed action should satisfy local state-level constraints. The first requirement supports safe policy learning; the second supports safe action selection through reach-avoid specifications, admissible action sets, or shielding.

At the policy level, we follow constrained reinforcement learning. For horizon $T$ and safety budget $d$, a cumulative safe-policy objective is
\begin{equation}
    \max_{\pi}\;
    \mathbb{E}_{\pi}\!\left[
        \sum_{t=0}^{T-1} r_{\kappa_t}(s_t,a_t)
    \right]
    \quad
    \text{s.t.}
    \quad
    \mathbb{E}_{\pi}\!\left[
        \sum_{t=0}^{T-1} c_{\kappa_t}(s_t,a_t)
    \right]\le d .
    \label{eq:cmdp_objective}
\end{equation}
A probabilistic safe-policy condition can also be written as
\begin{equation}
    \Pr_{\pi}\!\left(c_{\kappa_t}(s_t,a_t)>\tau_t\right)\le \delta_t,
    \qquad \forall t,
    \label{eq:probabilistic_safety}
\end{equation}
where $\tau_t$ is a context-dependent violation threshold and $\delta_t$ an allowable violation probability. These policy-level requirements can be implemented through constrained optimization, modified objectives, Lagrangian penalties, or reward shaping.

Cumulative or probabilistic safety requirements can affect decision-time behavior either directly, by inducing hard state-level constraints, or indirectly, by defining budget-type requirements that must be enforced over time. Context-based and adaptation-speed mechanisms may directly construct context-dependent reach-avoid specifications or exclude future state regions where safe adaptation is infeasible. When safety is represented as a cumulative cost limit or violation budget, we track the remaining safety budget
\begin{equation}
    B_t=d-\sum_{\tau=0}^{t-1}c_\tau,
    \qquad
    c_\tau=c_{\kappa_\tau}(s_\tau,a_\tau),
    \label{eq:remaining_budget}
\end{equation}
which measures the residual safety allowance at time $t$. This quantity enables soft-to-hard budget-driven constraints to convert cumulative requirements into local decision-time constraints.

\paragraph{Hard state-level safety.}
In safety-critical settings, cumulative or probabilistic safety alone may be insufficient. Certain actions must be ruled out at the current state if they violate a hard safety specification. We represent such a specification by
\begin{equation}
    h_t(s,a)\le 0,
    \label{eq:hard_constraint}
\end{equation}
with corresponding admissible action set
\begin{equation}
    \mathcal{A}^{\mathrm{hard}}_t(s)
    =
    \{a\in\mathcal{A}:h_t(s,a)\le 0\}.
    \label{eq:hard_safe_action_set}
\end{equation}
When hard enforcement is active, the executed action must satisfy
\begin{equation}
    \tilde{a}_t\in\mathcal{A}^{\mathrm{hard}}_t(s_t).
    \label{eq:hard_execution}
\end{equation}
The hard constraint may come from a context-based safety specification, an adaptation-speed assessment, a reach-avoid condition, a collision-avoidance rule, a physical limit, or a soft-to-hard conversion of cumulative safety requirements using $B_t$ and context.

\paragraph{Adaptation speed.}
Predicted context changes may require the agent to adapt within a limited time. Let $\Delta_{\mathcal{K}}$ be a discrepancy measure over contexts. We summarize required adaptation speed as
\begin{equation}
    v^{\mathrm{req}}_t
    =
    \frac{1}{H}\Delta_{\mathcal{K}}(\hat{\kappa}_{t+H},\hat{\kappa}_t),
    \label{eq:required_adaptation_speed}
\end{equation}
and compare it with the agent's achievable adaptation speed $v^{\mathrm{cap}}_t$:
\begin{equation}
    \rho_t=
    \frac{v^{\mathrm{req}}_t}{v^{\mathrm{cap}}_t+\epsilon},
    \qquad \epsilon>0.
    \label{eq:adaptation_ratio}
\end{equation}
When $\rho_t>1$, predicted nonstationarity may outpace the agent's ability to adapt safely.

\paragraph{Adaptive constraint construction.}
The two safety outputs can be generated from a common adaptive safety-constraint construction mechanism. At time $t$, the mechanism receives $\hat{\kappa}_t$, $\hat{\kappa}_{t+1:t+H}$, $B_t$, and $\rho_t$. It constructs policy-level safety terms
\begin{equation}
    \Gamma^{\pi}_t
    =
    \Gamma^{\pi}(\hat{\kappa}_t,\hat{\kappa}_{t+1:t+H},B_t,\rho_t),
    \label{eq:policy_constraint_terms}
\end{equation}
and state-level hard constraints
\begin{equation}
    h_t(s,a)=h(s,a;\hat{\kappa}_t,\hat{\kappa}_{t+1:t+H},B_t,\rho_t).
    \label{eq:adaptive_hard_constraint}
\end{equation}
The first object supports safe policy learning; the second supports safe action selection. The dependence on $B_t$ is necessary when hard constraints are obtained through soft-to-hard budget conversion, while context-based and adaptation-speed mechanisms may also generate hard constraints directly from context forecasts and adaptation-feasibility information.

\paragraph{Problem statement.}
Given a nonstationary environment with context $\kappa_t$, the goal is to learn a context-conditioned policy $\pi(a\mid s,\hat{\kappa})$ and an adaptive safety-constraint construction mechanism that jointly support:
\begin{enumerate}
    \item \emph{safe policy learning}: maximize cumulative reward while satisfying cumulative or probabilistic safety constraints of the form in Eq.~\eqref{eq:cmdp_objective} or Eq.~\eqref{eq:probabilistic_safety};
    \item \emph{hard decision-time enforcement}: construct state-level constraints or reach-avoid admissibility conditions of the form in Eq.~\eqref{eq:hard_constraint}, so that executed actions satisfy Eq.~\eqref{eq:hard_execution}.
\end{enumerate}
LILAC+ addresses this problem by using current and predicted context to construct three complementary adaptive constraint types: CB, AS, and SH constraints. Together, these components keep safety tied to stable high-level goals while allowing concrete constraints and enforcement rules to change with current and predicted nonstationarity.
\section{Related Work}

\paragraph{Safe reinforcement learning under nonstationarity.}
Safe reinforcement learning is commonly formulated through safety costs, constrained optimization, risk-sensitive objectives, and shielding~\citep{garcia2015comprehensive,gu2022review,chow2015risk,tamar2015policy,alshiekh2018safe}. Constrained Markov decision processes and constrained policy optimization provide foundations for policy-level safety through expected cumulative cost bounds~\citep{altman1999constrained,achiam2017constrained}, while shielding methods enforce action-level safety using predefined safe sets or safety specifications~\citep{alshiekh2018safe,jansen2020safe,carr2023safe}. Recent work extends safe policy improvement, constrained reinforcement learning, and online primal--dual optimization to nonstationary dynamics, rewards, or constraints~\citep{chandak2020towards,ding2023provably,wei2023provably,qiu2020upper}. These methods provide important foundations, but they often treat safety costs, constraints, or shields as fixed or externally specified rather than adaptive objects that change with the environment. LILAC+ instead constructs context-predictive adaptive constraints for both policy-level safety and action-level enforcement.

\paragraph{Context-aware, predictive, and adaptation-aware safety.}
Nonstationary reinforcement learning has also been studied through domain adaptation, meta-learning, lifelong learning, continual learning, and contextual reinforcement learning~\citep{brunskill2014pac,abel2018policy,parisi2019continual,xie2020deep,al2017continuous,chandak2020optimizing,chen2021context}. These approaches adapt policies to changing dynamics, rewards, tasks, or latent contexts, but often focus on reward performance rather than safety during context transitions. Prediction and uncertainty estimation have been used for safety in model-based safe reinforcement learning, Gaussian-process safety models, control barrier functions, and robust optimization~\citep{berkenkamp2015safe,berkenkamp2017safe,choi2020reinforcement,daulton2022robust}. Such methods help avoid unsafe regions, but are not always designed for continual nonstationarity in which concrete safety constraints themselves must change with context. LILAC+ differs by using detected and predicted context to coordinate CB, AS, and SH constraints within a shared predictive-context framework.
\section{Method}
\label{sec:method}

\subsection{LILAC+ overview}

LILAC+ constructs adaptive safety constraints for continual reinforcement learning under nonstationarity. At each time step, it receives $s_t$, detected context $\hat{\kappa}_t$, predicted context $\hat{\kappa}_{t+1:t+H}$, remaining budget $B_t$, and adaptation-speed ratio $\rho_t$. These inputs are used to construct three mechanisms: context-based constraints (CB), adaptation-speed constraints (AS), and soft-to-hard budget-driven constraints (SH).

The resulting constraints support two safety outputs. The \emph{safe policy} output shapes learning through modified objectives, cumulative constraints, or reward shaping. The \emph{safe action} output defines admissible action sets for reach-avoid specifications or runtime shielding.

\subsection{Shared predictive context pipeline}

LILAC+ uses the context detector $D$ and predictor $F$ defined in Section~\ref{sec:background} as a shared input layer for CB, AS, and SH constraints. At time $t$, the detector estimates the current context $\hat{\kappa}_t$ from the interaction history, and the predictor estimates future safety-relevant context $\hat{\kappa}_{t+1:t+H}$. The detected context is used by the context-conditioned policy $\pi(a\mid s,\hat{\kappa})$, while the predicted context allows constraints to be adapted before anticipated changes increase violation risk.

The predicted context sequence also defines the adaptation-speed ratio $\rho_t$ from Eq.~\eqref{eq:adaptation_ratio}. When $\rho_t>1$, predicted nonstationarity may require faster adaptation than the agent can safely achieve. This signal is used by AS constraints and may also influence CB and SH constraint construction.

\paragraph{Concrete instantiation.}
In the evaluated \texttt{merge-v0} domain, the context is the tuple $\kappa_t=(d_t,b_t,n_t)$ for traffic density, behavior/aggressiveness, and noise/sensing regime. The detector $D$ reads the simulator context and recent safety events from $\mathcal{H}_t$. The predictor $F$ is a finite-state transition model with persistence smoothing: it forecasts the most likely context sequence using empirical transition counts and a self-transition prior. The adaptation capacity $v_t^{\mathrm{cap}}$ is estimated from the largest recent context shift for which the safety layer restored admissible execution without fallback; if no such recovery is observed, a conservative default is used. The budget allocator is
\begin{equation}
    \phi(B_t,\hat{\kappa}_t,\hat{\kappa}_{t+1:t+H},\rho_t)
    =
    \frac{B_t}{N_t+\epsilon}
    \cdot
    \frac{1}{1+\alpha R_t+\beta\max(0,\rho_t-1)},
    \label{eq:concrete_phi}
\end{equation}
where $R_t$ is the predicted risk level derived from density, aggressiveness, and sensing noise. Thus high predicted risk or insufficient adaptation capacity reduces the local threshold.

\subsection{Adaptive constraint types}

\paragraph{CB constraints.}
CB constraints adapt safety requirements to the current and predicted context:
\begin{equation}
    g^{\mathrm{CB}}_t(s,a)
    =
    g^{\mathrm{CB}}
    (s,a;\hat{\kappa}_t,\hat{\kappa}_{t+1:t+H}).
    \label{eq:cb_constraint}
\end{equation}
They determine which safety requirements are appropriate under anticipated nonstationarity. In driving, for example, following distance, admissible acceleration, or collision-risk thresholds may become more conservative when predicted context indicates denser traffic or higher interaction risk. CB constraints primarily support immediate-to-mid-horizon safety adaptation.

\paragraph{AS constraints.}
AS constraints use $\rho_t$ to assess whether predicted changes can be safely tracked:
\begin{equation}
    g^{\mathrm{AS}}_t(s,a)
    =
    g^{\mathrm{AS}}
    (s,a;\hat{\kappa}_t,\hat{\kappa}_{t+1:t+H},\rho_t).
    \label{eq:as_constraint}
\end{equation}
If $\rho_t \le 1$, the required adaptation speed is estimated to be within the agent's adaptation capacity. If $\rho_t>1$, LILAC+ treats the associated future states, transitions, or actions more conservatively, for example by tightening thresholds, increasing safety penalties, or excluding actions that move the agent toward anticipated unsafe regions. AS constraints primarily support mid-to-long-horizon safety reasoning.

\paragraph{SH constraints.}
SH constraints convert cumulative or budget-type safety requirements into enforceable state-level constraints. Given remaining budget $B_t$, LILAC+ computes a local threshold
\begin{equation}
    \tau_t
    =
    \phi(B_t,\hat{\kappa}_t,\hat{\kappa}_{t+1:t+H},\rho_t),
    \label{eq:budget_threshold}
\end{equation}
where $\phi$ allocates remaining budget using current context, predicted context, and adaptation-speed information. A simple instance is
\begin{equation}
    \tau_t
    =
    \frac{B_t}{N_t+\epsilon},
    \label{eq:simple_budget_projection}
\end{equation}
where $N_t$ is the remaining horizon or estimated future exposure. The corresponding SH constraint is
\begin{equation}
    g^{\mathrm{SH}}_t(s,a)
    =
    c_{\hat{\kappa}_t}(s,a)-\tau_t .
    \label{eq:sh_constraint}
\end{equation}
This component connects cumulative safety requirements to decision-time enforcement by converting soft safety budgets into local state-level constraints.

\subsection{Safe policy and safe action outputs}

For policy-level safety, LILAC+ constructs safety terms
\begin{equation}
    \Gamma^{\pi}_t
    =
    \Gamma^{\pi}
    \big(
    g^{\mathrm{CB}}_t,
    g^{\mathrm{AS}}_t,
    g^{\mathrm{SH}}_t,
    B_t,
    \rho_t
    \big),
    \label{eq:policy_safety_terms}
\end{equation}
which may define cumulative constraints, probabilistic bounds, Lagrangian penalties, modified objectives, or reward-shaping terms. A generic penalized objective is
\begin{equation}
    \max_{\pi}
    \;
    \mathbb{E}_{\pi}
    \left[
        \sum_{t=0}^{T-1}
        r_{\kappa_t}(s_t,a_t)
        -
        \lambda_t
        \ell_t(s_t,a_t;\Gamma^{\pi}_t)
    \right],
    \label{eq:safe_policy_objective}
\end{equation}
where $\ell_t$ is a safety loss induced by the adaptive constraints and $\lambda_t \ge 0$ controls the strength of policy-level safety regularization.

For hard decision-time enforcement, LILAC+ combines the active constraints into
\begin{equation}
    h_t(s,a)
    =
    \max
    \left\{
        g^{\mathrm{CB}}_t(s,a),
        g^{\mathrm{AS}}_t(s,a),
        g^{\mathrm{SH}}_t(s,a)
    \right\},
    \label{eq:combined_hard_constraint}
\end{equation}
with admissible action set
\begin{equation}
    \mathcal{A}^{\mathrm{safe}}_t(s)
    =
    \{a \in \mathcal{A}: h_t(s,a) \le 0\}.
    \label{eq:method_safe_action_set}
\end{equation}
Given a proposed action $a_t^{\pi}\sim\pi(\cdot\mid s_t,\hat{\kappa}_t)$, the executed action is
\begin{equation}
    \tilde{a}_t
    =
    \begin{cases}
    a_t^{\pi},
    &
    \text{if } a_t^{\pi} \in \mathcal{A}^{\mathrm{safe}}_t(s_t),\\
    \Pi_{\mathcal{A}^{\mathrm{safe}}_t(s_t)}(a_t^{\pi}),
    &
    \text{otherwise,}
    \end{cases}
    \label{eq:shielded_action}
\end{equation}
where $\Pi_{\mathcal{A}^{\mathrm{safe}}_t(s_t)}$ denotes a projection, replacement, or fallback rule that selects an admissible action.

\subsection{Continual update and conditional safety}

After executing $\tilde{a}_t$, the agent observes reward $r_t$, cost $c_t$, and next state $s_{t+1}$. The remaining budget is updated as
\begin{equation}
    B_{t+1}=B_t-c_t,
    \qquad
    B_0=d.
    \label{eq:method_budget_update}
\end{equation}
The policy, context detector, and context predictor may then be updated using the new transition.

LILAC+ provides conditional safety properties under explicit assumptions rather than unconditional guarantees. If $\mathcal{A}^{\mathrm{safe}}_t(s_t)$ is non-empty and the enforcement rule in Eq.~\eqref{eq:shielded_action} is applied correctly, then $\tilde{a}_t\in\mathcal{A}^{\mathrm{safe}}_t(s_t)$ and therefore $h_t(s_t,\tilde{a}_t)\le 0$. If the SH threshold $\tau_t$ is a valid allocation of the remaining budget and $g^{\mathrm{SH}}_t(s_t,\tilde{a}_t)\le 0$, then $c_t\le\tau_t$, linking local enforcement to cumulative safety-budget control. These properties rely on sufficiently accurate context detection and prediction, valid CB/AS/SH constraint construction, bounded nonstationarity over the safety horizon, and a non-empty safe action set or feasible fallback action. Additional safety arguments and the full LILAC+ procedure are provided in ~\ref{app:safety_arguments} and~\ref{app:algorithm}, respectively.
\section{Experiments}
\label{sec:experiments}

\subsection{Experimental setup}
\label{sec:experimental_setup}

We evaluate LILAC+ in nonstationary reinforcement learning settings designed to test whether adaptive constraints reduce safety violations under context shift while preserving useful task performance. The experiments address three questions: (i) whether LILAC+ improves safety relative to unconstrained and fixed-constraint baselines, (ii) whether the safety improvement preserves competitive reward, and (iii) whether CB, AS, and SH constraints provide complementary benefits.

Experiments are conducted in \texttt{highway-env} using the \texttt{merge-v0} autonomous-driving task~\citep{leurent2018highwayenv}. The ego vehicle must merge onto a highway while interacting with surrounding traffic. Nonstationarity is induced by controlled changes in traffic density, driving aggressiveness, and sensor/noise conditions. We evaluate stationary, seen nonstationary, and unseen nonstationary conditions. The seen condition contains context changes represented during training, while the unseen condition contains held-out distribution shifts.

\paragraph{Implementation.}
We use the concrete instantiation in Section~\ref{sec:method}. CB maps context forecasts to front-gap, time-to-collision, merge-gap, and closing-speed thresholds; AS tightens these thresholds when $\rho_t>1$; and SH maintains $B_t$, allocates $\tau_t$ using Eq.~\eqref{eq:concrete_phi}, and invokes a conservative fallback when the admissible set is violated. The same learning backbone is used across baselines.

We compare against two baselines. The unconstrained baseline measures task performance without safety regulation. The fixed-constraint baseline applies a stationary safety specification uniformly across contexts. LILAC+ constructs adaptive constraints using current and predicted context, adaptation-speed information, and remaining safety budget. Violation count over the evaluation horizon is the primary safety metric and reward is the primary performance metric; tables report mean and standard deviation across random seeds. Additional experimental details are provided in ~\ref{app:experimental_details}.

\subsection{Results}
\label{sec:results}

The main evaluation compares full LILAC+ with fixed stationary constraints and unconstrained reinforcement learning under the strongest nonstationary condition. We also evaluate portability under an alternative learning backbone and perform ablations of CB, AS, SH, and their combinations.

\begin{figure}[t]
\centering

\begin{minipage}{0.48\linewidth}
    \centering
    \includegraphics[width=\linewidth]{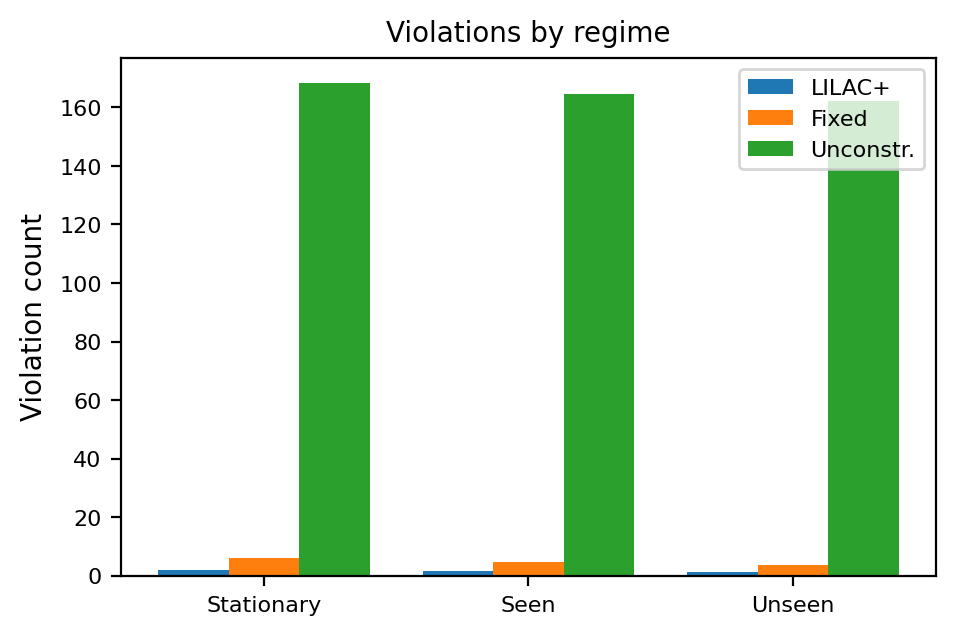}
    {\footnotesize (a) Main violations}
\end{minipage}
\hfill
\begin{minipage}{0.48\linewidth}
    \centering
    \includegraphics[width=\linewidth]{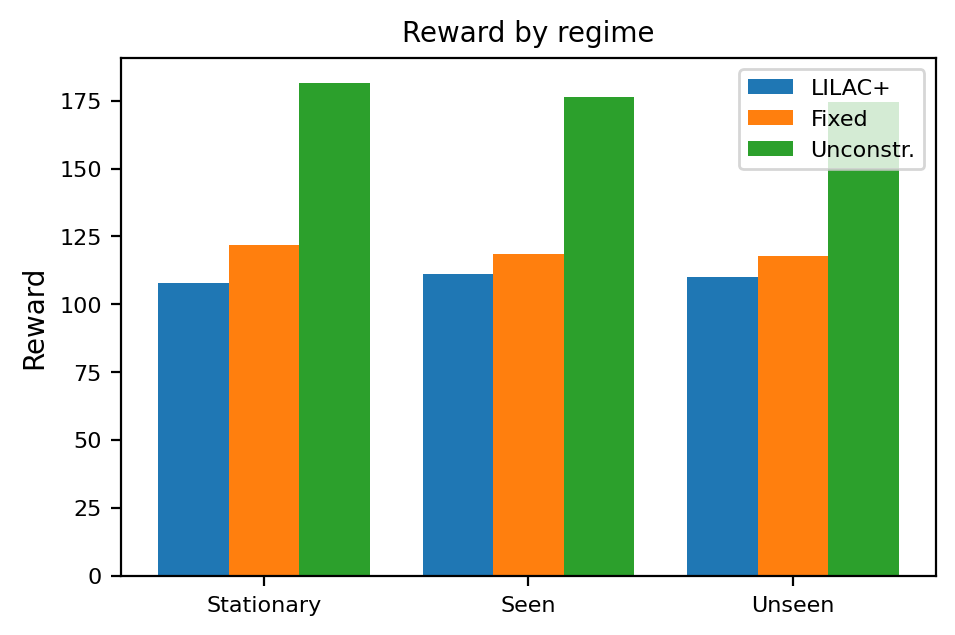}
    {\footnotesize (b) Reward}
\end{minipage}

\vspace{1mm}

\begin{minipage}{0.48\linewidth}
    \centering
    \includegraphics[width=\linewidth]{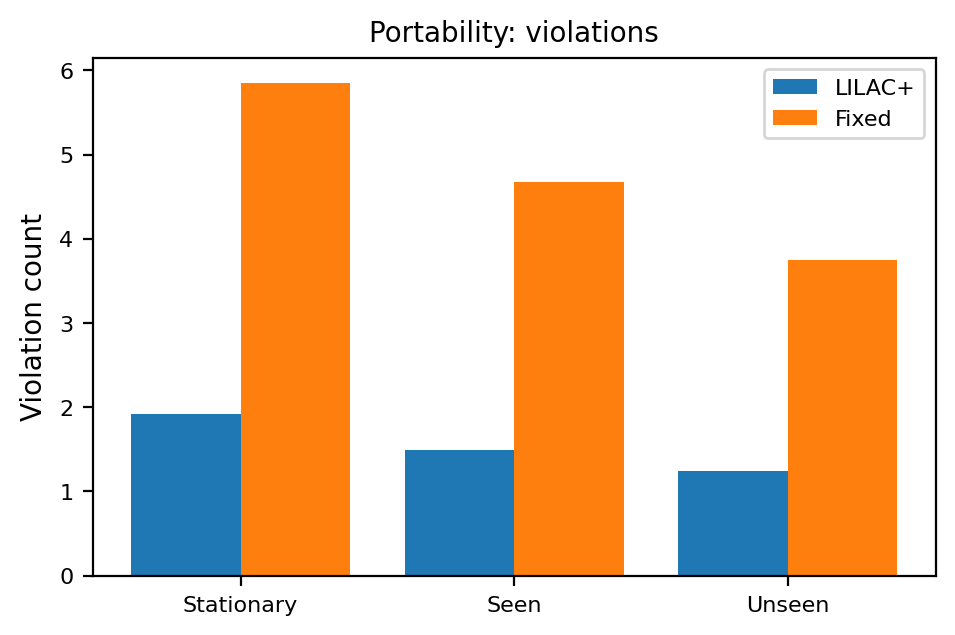}
    {\footnotesize (c) Portability}
\end{minipage}
\hfill
\begin{minipage}{0.48\linewidth}
    \centering
    \includegraphics[width=\linewidth]{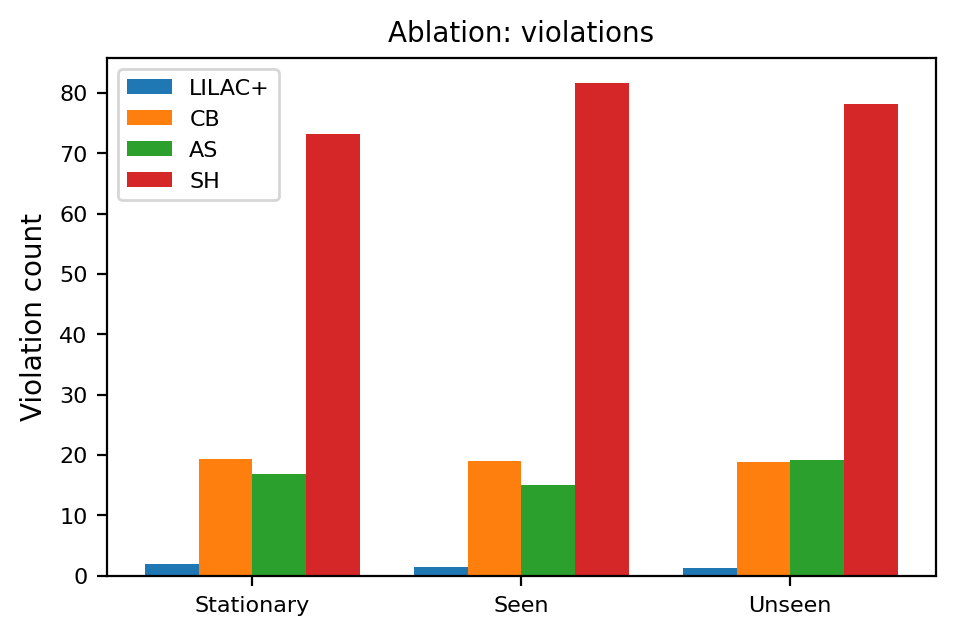}
    {\footnotesize (d) Ablation}
\end{minipage}

\caption{
Experimental evidence for adaptive safety constraints. LILAC+ reduces violations in the main nonstationary comparison, preserves useful reward, remains effective with an alternative learning backbone, and outperforms partial CB/AS/SH variants.
}
\label{fig:experimental_evidence}
\end{figure}

Table~\ref{tab:D11_C8_summary} summarizes the main nonstationary comparison, and Figure~\ref{fig:experimental_evidence} summarizes safety, reward, portability, and ablation trends.
\begin{table}[H]
\centering
\small
\caption{
Main nonstationary comparison. Values are reported as mean $\pm$ standard deviation over 30 runs from 10 random seeds. Violation count is the primary safety metric, reward measures task performance, and clearance is interpreted as a safety buffer; lower violation at moderate clearance indicates greater safety efficiency.
}
\label{tab:D11_C8_summary}
\begin{tabular}{lccc}
\toprule
Method & Violation Count $\downarrow$ & Reward $\uparrow$ & Clearance \\
\midrule
Unconstrained RL
& $165.10 \pm 15.05$
& $177.56 \pm 8.87$
& $1857.36 \pm 155.09$ \\

Fixed stationary constraints
& $4.76 \pm 0.09$
& $119.48 \pm 2.74$
& $2024.26 \pm 43.12$ \\

Full adaptive constraints (CB+AS+SH)
& $1.55 \pm 0.01$
& $109.70 \pm 1.21$
& $1913.49 \pm 30.03$ \\
\bottomrule
\end{tabular}
\end{table}

The results show that unconstrained reinforcement learning obtains high reward but incurs substantially more violations. Fixed stationary constraints reduce violations, but remain vulnerable under strong nonstationarity. LILAC+ achieves the lowest violation count by adapting constraints using current and predicted context, adaptation-speed information, and remaining safety budget. This comes with some reward reduction relative to fixed constraints, but the reward remains competitive given the large safety improvement.

The ablation results indicate that the three adaptive constraint types provide complementary benefits. CB supports immediate-to-mid-horizon adjustment of safety requirements, AS supports mid-to-long-horizon safety assessment by tightening behavior when the speed required for safe adaptation exceeds the agent's estimated adaptation speed, and SH strengthens decision-time enforcement by converting cumulative safety requirements into local state-level constraints. The full CB+AS+SH configuration yields the lowest violation counts, suggesting that safety under nonstationarity benefits from combining context-based adaptation, adaptation-speed assessment, and budget-aware enforcement.
\subsection{Scope}
\label{sec:experimental_scope}

These experiments are intentionally limited to simulated \texttt{merge-v0} driving with controlled context shifts. They support the proposed CB/AS/SH mechanisms in this setting, but should not be read as universal safety guarantees. Broader validation requires additional domains, stronger safe-RL baselines, richer nonstationarity, and deeper analysis of prediction error and fallback feasibility.

\section{Conclusion}

We presented LILAC+, a predictive-context framework for safe continual reinforcement learning under nonstationarity. The central idea is to treat safety as context-dependent constraint adaptation: high-level safety goals may remain stable, but the concrete constraints used to implement them should change with current and predicted context. LILAC+ uses context detection and context prediction to construct CB, AS, and SH constraints, supporting both safe policy learning through modified objectives, cumulative constraints, or probabilistic bounds, and safe action selection through state-level constraints, reach-avoid specifications, or runtime shielding.

Experiments in nonstationary highway-driving settings indicate that LILAC+ reduces safety violations and cumulative safety cost relative to unconstrained and fixed-constraint baselines while maintaining competitive reward. These results suggest that predictive adaptive constraints are a practical mechanism for improving safety in continual reinforcement learning under changing environmental conditions.

\section{Limitations}

LILAC+ depends on the quality of context detection and context prediction. If the detected context is inaccurate, or if predicted context shifts fall outside the predictor's support, the constructed constraints may be misaligned with the true safety requirements. This is especially important under abrupt or previously unseen nonstationarity.

The conditional safety properties also rely on valid constraint construction and feasible enforcement. In particular, the soft-to-hard budget-driven constraint requires a meaningful conversion from remaining safety budget to local state-level thresholds, and runtime enforcement requires a non-empty admissible action set or feasible fallback action. When these assumptions fail, strict decision-time safety cannot be guaranteed.

Finally, the present empirical evaluation is limited to simulated highway-driving tasks with controlled context shifts. Broader validation is needed in additional domains, with richer forms of nonstationarity, more realistic perception uncertainty, and tighter analysis of context-prediction error, fallback feasibility, and computational overhead.

\bibliographystyle{plainnat}
\bibliography{bib}

@book{altman1999constrained,
  title     = {Constrained Markov decision processes},
  author    = {Altman, Eitan},
  volume    = {7},
  year      = {1999},
  publisher = {CRC press}
}

@article{choi2020reinforcement,
  title   = {Reinforcement learning for safety-critical control under model uncertainty, using control lyapunov functions and control barrier functions},
  author  = {Choi, Jason and Castaneda, Fernando and Tomlin, Claire J and Sreenath, Koushil},
  journal = {arXiv preprint arXiv:2004.07584},
  year    = {2020}
}

@article{gu2022review,
  title   = {A review of safe reinforcement learning: Methods, theory and applications},
  author  = {Gu, Shangding and Yang, Long and Du, Yali and Chen, Guang and Walter, Florian and Wang, Jun and Yang, Yaodong and Knoll, Alois},
  journal = {arXiv preprint arXiv:2205.10330},
  year    = {2022}
}

@article{xie2020deep,
  title   = {Deep reinforcement learning amidst lifelong non-stationarity},
  author  = {Xie, Annie and Harrison, James and Finn, Chelsea},
  journal = {arXiv preprint arXiv:2006.10701},
  year    = {2020}
}

@inproceedings{achiam2017constrained,
  title     = {Constrained Policy Optimization},
  author    = {Achiam, Joshua and Held, David and Tamar, Aviv and Abbeel, Pieter},
  booktitle = {International Conference on Machine Learning (ICML)},
  year      = {2017}
}

@inproceedings{chow2015risk,
  title     = {Risk-Sensitive and Robust Decision-Making: a {CVaR} Optimization Approach},
  author    = {Chow, Yinlam and Ghavamzadeh, Mohammad and Janson, Lucas and Pavone, Marco},
  booktitle = {Advances in Neural Information Processing Systems (NeurIPS)},
  year      = {2015}
}

@article{tamar2015policy,
  title   = {Policy Gradient for Coherent Risk Measures},
  author  = {Tamar, Aviv and Glassner, Yonatan and Mannor, Shie},
  journal = {arXiv preprint arXiv:1502.02267},
  year    = {2015}
}

@inproceedings{berkenkamp2017safe,
  title     = {Safe Model-based Reinforcement Learning with Stability Guarantees},
  author    = {Berkenkamp, Felix and Turchetta, Marcello and Schoellig, Angela P. and Krause, Andreas},
  booktitle = {Advances in Neural Information Processing Systems (NeurIPS)},
  year      = {2017}
}

@inproceedings{berkenkamp2015safe,
  title        = {Safe and robust learning control with Gaussian processes},
  author       = {Berkenkamp, Felix and Schoellig, Angela P},
  booktitle    = {2015 European Control Conference (ECC)},
  pages        = {2496--2501},
  year         = {2015},
  organization = {IEEE}
}

@inproceedings{daulton2022robust,
  title        = {Robust multi-objective bayesian optimization under input noise},
  author       = {Daulton, Samuel and Cakmak, Sait and Balandat, Maximilian and Osborne, Michael A and Zhou, Enlu and Bakshy, Eytan},
  booktitle    = {International Conference on Machine Learning},
  pages        = {4831--4866},
  year         = {2022},
  organization = {PMLR}
}

@article{parisi2019continual,
  title     = {Continual lifelong learning with neural networks: A review},
  author    = {Parisi, German I and Kemker, Ronald and Part, Jose L and Kanan, Christopher and Wermter, Stefan},
  journal   = {Neural networks},
  volume    = {113},
  pages     = {54--71},
  year      = {2019},
  publisher = {Elsevier}
}

@inproceedings{brunskill2014pac,
  title        = {Pac-inspired option discovery in lifelong reinforcement learning},
  author       = {Brunskill, Emma and Li, Lihong},
  booktitle    = {International conference on machine learning},
  pages        = {316--324},
  year         = {2014},
  organization = {PMLR}
}

@inproceedings{abel2018policy,
  title        = {Policy and value transfer in lifelong reinforcement learning},
  author       = {Abel, David and Jinnai, Yuu and Guo, Sophie Yue and Konidaris, George and Littman, Michael},
  booktitle    = {International conference on machine learning},
  pages        = {20--29},
  year         = {2018},
  organization = {PMLR}
}

@article{al2017continuous,
  title   = {Continuous adaptation via meta-learning in nonstationary and competitive environments},
  author  = {Al-Shedivat, Maruan and Bansal, Trapit and Burda, Yuri and Sutskever, Ilya and Mordatch, Igor and Abbeel, Pieter},
  journal = {arXiv preprint arXiv:1710.03641},
  year    = {2017}
}

@inproceedings{ding2023provably,
  title     = {Provably efficient primal-dual reinforcement learning for cmdps with non-stationary objectives and constraints},
  author    = {Ding, Yuhao and Lavaei, Javad},
  booktitle = {Proceedings of the AAAI Conference on Artificial Intelligence},
  volume    = {37},
  pages     = {7396--7404},
  year      = {2023}
}

@article{chandak2020towards,
  title   = {Towards safe policy improvement for non-stationary MDPs},
  author  = {Chandak, Yash and Jordan, Scott and Theocharous, Georgios and White, Martha and Thomas, Philip S},
  journal = {Advances in Neural Information Processing Systems},
  volume  = {33},
  pages   = {9156--9168},
  year    = {2020}
}

@article{qiu2020upper,
  title   = {Upper confidence primal-dual optimization: Stochastically constrained markov decision processes with adversarial losses and unknown transitions},
  author  = {Qiu, Shuang and Wei, Xiaohan and Yang, Zhuoran and Ye, Jieping and Wang, Zhaoran},
  journal = {arXiv preprint arXiv:2003.00660},
  year    = {2020}
}

@inproceedings{chandak2020optimizing,
  title        = {Optimizing for the future in non-stationary mdps},
  author       = {Chandak, Yash and Theocharous, Georgios and Shankar, Shiv and White, Martha and Mahadevan, Sridhar and Thomas, Philip},
  booktitle    = {International Conference on Machine Learning},
  pages        = {1414--1425},
  year         = {2020},
  organization = {PMLR}
}

@inproceedings{wei2023provably,
  title        = {Provably efficient model-free algorithms for non-stationary cmdps},
  author       = {Wei, Honghao and Ghosh, Arnob and Shroff, Ness and Ying, Lei and Zhou, Xingyu},
  booktitle    = {International Conference on Artificial Intelligence and Statistics},
  pages        = {6527--6570},
  year         = {2023},
  organization = {PMLR}
}

@inproceedings{chen2021context,
  title        = {Context-aware safe reinforcement learning for non-stationary environments},
  author       = {Chen, Baiming and Liu, Zuxin and Zhu, Jiacheng and Xu, Mengdi and Ding, Wenhao and Li, Liang and Zhao, Ding},
  booktitle    = {2021 IEEE International Conference on Robotics and Automation (ICRA)},
  pages        = {10689--10695},
  year         = {2021},
  organization = {IEEE}
}

@inproceedings{carr2023safe,
  title     = {Safe reinforcement learning via shielding under partial observability},
  author    = {Carr, Steven and Jansen, Nils and Junges, Sebastian and Topcu, Ufuk},
  booktitle = {Proceedings of the AAAI conference on artificial intelligence},
  volume    = {37},
  number    = {12},
  pages     = {14748--14756},
  year      = {2023}
}

@article{garcia2015comprehensive,
  title={A comprehensive survey on safe reinforcement learning},
  author={Garcia, Javier and Fernandez, Fernando},
  journal={JMLR},
  year={2015}
}

@inproceedings{alshiekh2018safe,
  title={Safe Reinforcement Learning via Shielding},
  author={Alshiekh, Mohammed and Bloem, Roderick and Ehlers, Ruediger and Könighofer, Bettina and Niekum, Scott and Topcu, Ufuk},
  booktitle={AAAI},
  year={2018}
}

@misc{leurent2018highwayenv,
  author       = {Leurent, Edouard},
  title        = {An Environment for Autonomous Driving Decision-Making},
  year         = {2018},
  howpublished = {\url{https://github.com/Farama-Foundation/HighwayEnv}},
  note         = {Highway-env}
}

@article{raffin2021stable,
  author  = {Raffin, Antonin and Hill, Ashley and Gleave, Adam and Kanervisto, Anssi and Ernestus, Maximilian and Dormann, Noah},
  title   = {Stable-Baselines3: Reliable Reinforcement Learning Implementations},
  journal = {Journal of Machine Learning Research},
  volume  = {22},
  number  = {268},
  pages   = {1--8},
  year    = {2021}
}

@inproceedings{jansen2020safe,
  author    = {Jansen, Nils and K{\"o}nighofer, Bettina and Junges, Sebastian and Serban, Alex and Bloem, Roderick},
  title     = {Safe Reinforcement Learning Using Probabilistic Shields},
  booktitle = {CONCUR},
  pages     = {3:1--3:16},
  year      = {2020},
  doi       = {10.4230/LIPIcs.CONCUR.2020.3}
}
\clearpage
\appendix
\renewcommand{\thesection}{Appendix~\Alph{section}}
\section{Additional Safety Arguments}
\label{app:safety_arguments}

This appendix provides additional details for the conditional safety properties discussed in Section~\ref{sec:method}. The arguments are conditional on the correctness of the constructed constraints, the quality of context detection and prediction, the validity of budget-to-threshold conversion, and the existence of feasible safe actions. They should not be interpreted as unconditional safety guarantees.

\paragraph{Notation.}
At time $t$, LILAC+ detects the current context $\hat{\kappa}_t$, predicts future context $\hat{\kappa}_{t+1:t+H}$, tracks the remaining safety budget $B_t$, and computes the adaptation-speed ratio $\rho_t$. These quantities parameterize three adaptive constraint types:
\[
    g^{\mathrm{CB}}_t(s,a),
    \qquad
    g^{\mathrm{AS}}_t(s,a),
    \qquad
    g^{\mathrm{SH}}_t(s,a).
\]
For hard decision-time enforcement, the active state-level constraint is
\[
    h_t(s,a)
    =
    \max
    \left\{
        g^{\mathrm{CB}}_t(s,a),
        g^{\mathrm{AS}}_t(s,a),
        g^{\mathrm{SH}}_t(s,a)
    \right\},
\]
and the admissible action set is
\[
    \mathcal{A}^{\mathrm{safe}}_t(s)
    =
    \{a \in \mathcal{A}: h_t(s,a) \le 0\}.
\]
Equivalently, an action is admissible only if it satisfies all active context-based, adaptation-speed, and soft-to-hard budget-driven constraints.

\paragraph{Assumptions.}
The conditional safety arguments rely on the following assumptions.

\begin{enumerate}
    \item \textbf{Context accuracy.} The detected context $\hat{\kappa}_t$ and predicted context $\hat{\kappa}_{t+1:t+H}$ are sufficiently accurate over the safety-relevant horizon.
    \item \textbf{Valid constraint construction.} The constructed CB, AS, and SH constraints correctly encode the intended safety specification under the detected and predicted context.
    \item \textbf{Valid budget projection.} The local threshold $\tau_t$ used by the SH constraint is a valid allocation of the remaining safety budget $B_t$.
    \item \textbf{Feasible enforcement.} The admissible action set $\mathcal{A}^{\mathrm{safe}}_t(s_t)$ is non-empty, or a feasible fallback action is available.
    \item \textbf{Bounded nonstationarity.} Context changes are sufficiently bounded over the prediction horizon so that predicted context remains informative for constraint construction.
\end{enumerate}

\paragraph{Per-step admissibility.}
Suppose that $\mathcal{A}^{\mathrm{safe}}_t(s_t)$ is non-empty and that the runtime enforcement rule is applied correctly. Given a proposed action
\[
    a_t^{\pi} \sim \pi(\cdot \mid s_t,\hat{\kappa}_t),
\]
LILAC+ executes
\[
    \tilde{a}_t
    =
    \begin{cases}
    a_t^{\pi}, & \text{if } a_t^{\pi} \in \mathcal{A}^{\mathrm{safe}}_t(s_t),\\
    \Pi_{\mathcal{A}^{\mathrm{safe}}_t(s_t)}(a_t^{\pi}), & \text{otherwise,}
    \end{cases}
\]
where $\Pi_{\mathcal{A}^{\mathrm{safe}}_t(s_t)}$ denotes a projection, replacement, or fallback rule that returns an admissible action. Therefore,
\[
    \tilde{a}_t \in \mathcal{A}^{\mathrm{safe}}_t(s_t),
\]
and by construction,
\[
    h_t(s_t,\tilde{a}_t) \le 0.
\]
Since $h_t$ is the maximum of the active CB, AS, and SH constraints, this implies
\[
    g^{\mathrm{CB}}_t(s_t,\tilde{a}_t) \le 0,
    \qquad
    g^{\mathrm{AS}}_t(s_t,\tilde{a}_t) \le 0,
    \qquad
    g^{\mathrm{SH}}_t(s_t,\tilde{a}_t) \le 0.
\]
Thus, under the stated assumptions, the executed action satisfies all active decision-time constraints.

\paragraph{Cumulative budget consistency.}
The remaining budget evolves as
\[
    B_{t+1}=B_t-c_t,
    \qquad
    B_0=d.
\]
The SH constraint converts the remaining safety budget into a local decision-time threshold,
\[
    \tau_t
    =
    \phi(B_t,\hat{\kappa}_t,\hat{\kappa}_{t+1:t+H},\rho_t),
\]
and defines
\[
    g^{\mathrm{SH}}_t(s,a)
    =
    c_{\hat{\kappa}_t}(s,a)-\tau_t.
\]
If the executed action satisfies
\[
    g^{\mathrm{SH}}_t(s_t,\tilde{a}_t)\le 0,
\]
then
\[
    c_t \le \tau_t.
\]
Therefore, when $\tau_t$ is a valid allocation of the remaining budget, decision-time enforcement controls realized cost according to the remaining-budget dynamics. This provides the connection between cumulative policy-level safety and local state-level enforcement.

\paragraph{Adaptation-speed conservatism.}
The adaptation-speed ratio
\[
    \rho_t
    =
    \frac{v^{\mathrm{req}}_t}{v^{\mathrm{cap}}_t+\epsilon}
\]
compares the adaptation speed required by predicted nonstationarity with the agent's achievable adaptation speed. When $\rho_t>1$, predicted context change may outpace the agent's ability to adapt safely. In this case, AS constraints can tighten admissible actions, increase policy-level safety penalties, or mark anticipated future states or transitions as unsafe for entry. This does not guarantee safety under arbitrary nonstationarity, but it provides a conservative response when predicted context change exceeds the agent's estimated adaptation capacity.

\paragraph{Effect of context prediction.}
Context prediction does not merely update a fixed constraint set. Instead, it parameterizes the construction of all three adaptive constraint types. The predicted context $\hat{\kappa}_{t+1:t+H}$ affects CB constraints by indicating which safety requirements are appropriate under anticipated context changes, affects AS constraints by determining the required adaptation speed, and affects SH constraints by influencing how remaining budget should be allocated to future decisions. When predicted risk, adaptation-speed pressure, or budget scarcity increases, the admissible set
\[
    \mathcal{A}^{\mathrm{safe}}_t(s)
\]
is tightened through the combined constraint
\[
    h_t(s,a)
    =
    \max
    \left\{
        g^{\mathrm{CB}}_t(s,a),
        g^{\mathrm{AS}}_t(s,a),
        g^{\mathrm{SH}}_t(s,a)
    \right\}.
\]
This reduces exposure to unsafe actions under changing contexts, provided that context detection, context prediction, and constraint construction remain sufficiently accurate.

\paragraph{Safe policy interpretation.}
The same adaptive constraints can also support policy-level safety. LILAC+ may construct policy-level safety terms
\[
    \Gamma^{\pi}_t
    =
    \Gamma^{\pi}
    \big(
    g^{\mathrm{CB}}_t,
    g^{\mathrm{AS}}_t,
    g^{\mathrm{SH}}_t,
    B_t,
    \rho_t
    \big),
\]
which can be used in cumulative constraints, probabilistic safety bounds, Lagrangian penalties, modified optimization criteria, or reward-shaping terms. This corresponds to the safe-policy output of the framework. The hard admissible set $\mathcal{A}^{\mathrm{safe}}_t(s)$ corresponds to the safe-action output.

\paragraph{Failure modes.}
The conditional properties above may fail if context prediction is inaccurate, if nonstationarity changes abruptly outside the predictor's support, if the constructed constraints do not correctly represent the intended safety specification, if the budget-to-threshold conversion is invalid, or if no feasible fallback action exists. In these cases, LILAC+ may still act conservatively, but strict decision-time enforcement cannot be guaranteed.

\section{Full LILAC+ Procedure}
\label{app:algorithm}

Algorithm~\ref{alg:lilac_plus_appendix} summarizes the full procedure used by LILAC+.

\begin{algorithm}[H]
\caption{LILAC+: Predictive Adaptive Safety Constraints for Continual RL}
\label{alg:lilac_plus_appendix}
\begin{algorithmic}[1]
\STATE Initialize policy $\pi$, safety budget $B_0=d$, context detector $D$, and context predictor $F$
\FOR{$t=0,1,2,\dots$}
    \STATE Observe state $s_t$ and update history $\mathcal{H}_t$
    \STATE Detect current context $\hat{\kappa}_t = D(\mathcal{H}_t)$
    \STATE Predict future context $\hat{\kappa}_{t+1:t+H}=F(\mathcal{H}_t,\hat{\kappa}_t)$
    \STATE Compute required adaptation speed $v^{\mathrm{req}}_t$
    \STATE Estimate achievable adaptation speed $v^{\mathrm{cap}}_t$
    \STATE Compute adaptation-speed ratio $\rho_t = v^{\mathrm{req}}_t/(v^{\mathrm{cap}}_t+\epsilon)$
    \STATE Construct context-based constraint $g^{\mathrm{CB}}_t$
    \STATE Construct adaptation-speed constraint $g^{\mathrm{AS}}_t$
    \STATE Project remaining budget $B_t$ into local threshold $\tau_t$
    \STATE Construct soft-to-hard budget-driven constraint $g^{\mathrm{SH}}_t$
    \STATE Construct policy-level safety terms $\Gamma^{\pi}_t$
    \STATE Define hard constraint $h_t$ and admissible set $\mathcal{A}^{\mathrm{safe}}_t(s_t)$
    \STATE Sample proposed action $a_t^{\pi}\sim\pi(\cdot\mid s_t,\hat{\kappa}_t)$
    \IF{$a_t^{\pi}\in\mathcal{A}^{\mathrm{safe}}_t(s_t)$}
        \STATE Execute $\tilde{a}_t=a_t^{\pi}$
    \ELSE
        \STATE Execute fallback action $\tilde{a}_t=\Pi_{\mathcal{A}^{\mathrm{safe}}_t(s_t)}(a_t^{\pi})$
    \ENDIF
    \STATE Observe reward $r_t$, cost $c_t$, and next state $s_{t+1}$
    \STATE Update remaining budget $B_{t+1}=B_t-c_t$
    \STATE Update policy $\pi$ using reward and policy-level safety terms $\Gamma^{\pi}_t$
    \STATE Update context detector $D$ and context predictor $F$
\ENDFOR
\end{algorithmic}
\end{algorithm}

\section{Experimental Details}
\label{app:experimental_details}

\paragraph{Environment.}
Experiments use the \texttt{highway-env} \texttt{merge-v0} environment~\citep{leurent2018highwayenv}, an autonomous-driving reinforcement learning simulator in which an ego vehicle merges onto a highway while interacting with surrounding traffic. All experimental data are generated by simulation.

\paragraph{Nonstationarity.}
Nonstationarity is induced through controlled changes in context variables such as traffic density, relative vehicle speed, and inter-vehicle distance variability. These variables affect both task difficulty and safety requirements. The purpose of the nonstationarity protocol is to test whether safety constraints remain effective when operating conditions change over time.

\paragraph{Concrete LILAC+ implementation.}
For the reported \texttt{merge-v0} experiments, the detected context is $\hat{\kappa}_t=(d_t,b_t,n_t)$, where $d_t$ encodes traffic density, $b_t$ encodes behavior/aggressiveness, and $n_t$ encodes sensing/noise level. The detector $D$ reads the active simulator context and augments it with recent safety events in $\mathcal{H}_t$. The predictor $F$ is a finite-state model estimated from empirical context transitions with persistence smoothing, so the predicted sequence favors self-transitions unless recent observations support a context change.

The adaptation capacity $v_t^{\mathrm{cap}}$ is estimated from recent successful recoveries: it is the largest recent context discrepancy for which the safety layer returned to admissible execution without invoking fallback. If no recovery evidence is available, we use a conservative default. The allocator $\phi$ uses Eq.~\eqref{eq:concrete_phi}; predicted risk $R_t$ increases with density, aggressiveness, and sensing noise, while the factor $\max(0,\rho_t-1)$ tightens the threshold only when the required adaptation speed exceeds the estimated capacity. This supports the concrete definitions used by CB, AS, and SH in the main experiments.

\paragraph{Evaluation conditions.}
We evaluate three test conditions.
\begin{itemize}
    \item \textbf{Stationary condition:} context variables remain fixed during evaluation.
    \item \textbf{Seen nonstationary condition:} context changes follow patterns represented during training.
    \item \textbf{Unseen nonstationary condition:} context changes include held-out distribution shifts not observed during training.
\end{itemize}
This separation evaluates adaptation to known context variation and robustness to previously unseen nonstationary conditions.

\paragraph{Methods and baselines.}
The experiments compare LILAC+ with two primary baselines.
\begin{itemize}
    \item \textbf{Unconstrained RL:} a reinforcement learning agent trained without safety regulation.
    \item \textbf{Fixed stationary constraints:} a safety-constrained agent using a fixed safety specification applied uniformly across contexts.
    \item \textbf{LILAC+:} the proposed CB/AS/SH mechanism suite combining context-based constraints, adaptation-speed constraints, and soft-to-hard budget-driven constraints.
\end{itemize}
The underlying reinforcement learning implementation uses Stable-Baselines3~\citep{raffin2021stable}. The adaptive constraints are implemented as a safety layer around the learning backbone, enabling comparison across policy optimizers.

\paragraph{Reported tests.}
The main paper reports three experimental groups.
\begin{itemize}
    \item \textbf{Test C8: strong nonstationarity.} This is the main comparison between unconstrained RL, fixed stationary constraints, and full LILAC+ under strong context shifts.
    \item \textbf{Test C7: portability.} This evaluates whether adaptive constraints remain effective when used with an alternative learning backbone.
    \item \textbf{Test C5: ablation.} This evaluates partial combinations of CB, AS, and SH constraints to assess component complementarity.
\end{itemize}

\paragraph{Ablation variants.}
The ablation study evaluates whether the three adaptive constraint types contribute complementary benefits. Depending on the test configuration, partial variants may include CB-only, AS-only, SH-only, pairwise combinations, and the full CB+AS+SH configuration. CB tests context-based adaptation of safety requirements, AS tests adaptation-speed-aware tightening or avoidance, and SH tests conversion of cumulative safety requirements into local state-level constraints.

\paragraph{Metrics.}
We report violation count as the primary safety metric and reward as the primary performance metric. Violation count is computed as the number of evaluation time steps in which a safety violation occurs over the evaluation horizon. Reward measures task performance. We also report cumulative safety cost and clearance when relevant. Clearance is interpreted as a safety buffer rather than a standalone safety metric: low violation counts at moderate clearance indicate safety-efficient behavior, while very large clearance may indicate overly conservative behavior.

\paragraph{Random seeds and runs.}
Reported results are aggregated over 10 random seeds and 30 evaluation runs unless otherwise specified. Tables report mean $\pm$ standard deviation across runs.

\paragraph{Statistical reporting.}
For metric values $x_1,\ldots,x_n$, we report
\[
    \bar{x}
    =
    \frac{1}{n}\sum_{i=1}^{n}x_i,
    \qquad
    s
    =
    \sqrt{
    \frac{1}{n-1}
    \sum_{i=1}^{n}(x_i-\bar{x})^2
    }.
\]
The main results table reports mean $\pm$ standard deviation for violation count, reward, and clearance. Figures provide aggregate trends across stationary, seen nonstationary, and unseen nonstationary conditions.

\paragraph{Compute resources.}
All reported experiments were run on CPU. The reported experimental suite required approximately 750 total CPU-hours, including the main C8 comparison, C7 portability evaluation, and C5 ablation runs.

\paragraph{Reproducibility.}
An supplementary ZIP archive containing implementation, configuration files, and reproduction instructions is provided with the submission. The archive includes scripts for the proposed method and baselines, along with configuration files for the reported C5, C7, and C8 experiments.

\paragraph{Existing assets and licenses.}
Experiments use \texttt{highway-env} \texttt{merge-v0}~\citep{leurent2018highwayenv} and Stable-Baselines3~\citep{raffin2021stable}. We cite these open-source assets and respect their corresponding licenses and terms of use. Version and license details are provided in the code supplement.

\end{document}